\documentclass{article}

\usepackage{arxiv}

\usepackage[utf8]{inputenc} 
\usepackage[T1]{fontenc}    
\usepackage{hyperref}       
\usepackage{url}            
\usepackage{booktabs}       
\usepackage{amsfonts}       
\usepackage{nicefrac}       
\usepackage{microtype}      
\usepackage{lipsum}
\usepackage{graphicx}
\graphicspath{ {./images/} }
\usepackage{natbib}
\usepackage{amsmath}
\usepackage{caption}
\usepackage{algorithm}
\usepackage{algpseudocode}
\usepackage{multirow}

\title{Graph Classification via Network Usable Information: From Representation Evaluation to Structure Selection}

\author{
 Abdullah Shaik \\
  Cyber Security Consultant \\
  Regulatory Risk \& Forensic \\
  Deloitte Transaction and Business LLP \\
  Nashville, 37076 TN \\
  \texttt{sabdullah@deloitte.com} \\
   \And
 Anwar Said \\
  Research Scientist\\
  Vanderbilt University\\
  Nashville, 37212 TN \\
  \texttt{anwar.said@vanderbilt.edu} \\
  }

\begin{document}
\date{\vspace{-2em}}
\maketitle

\begin{abstract}
We propose \textsc{NetinfoGC}, a framework for graph classification that extends the Network Usable Information (NUI) paradigm to graph-level learning. Unlike conventional graph neural network approaches that rely on end-to-end training of black-box embeddings, \textsc{NetinfoGC} constructs a family of permutation-invariant graph representations derived from propagation-based mechanisms and classical structural descriptors, including graph centrality measures.

To evaluate representation quality, we introduce a training-free NUI estimation procedure based on clustering consistency with ground-truth labels, providing a proxy for task-relevant information without supervised learning. We further exploit the same representations using sparse-group LASSO regularization, enabling automatic selection of informative structural descriptors while suppressing redundant ones.

Experiments on benchmark datasets show that classical centrality measures are highly competitive with learned propagation-based representations, and in several cases yield superior performance. Moreover, we observe a strong correlation between estimated NUI and downstream classification accuracy, validating NUI as an effective measure of representation utility.

Overall, \textsc{NetinfoGC} provides a unified and interpretable framework for evaluating and exploiting graph representations without requiring end-to-end neural training.
\end{abstract}


\section{Introduction}
\label{sec:intro}

Graph classification is a fundamental problem in graph machine learning, where the objective is to assign labels to entire graphs based on their structural and attribute information. It arises in a wide range of applications, including molecular property prediction, program analysis, malware detection, and network security. In recent years, graph neural networks (GNNs) have become the dominant paradigm for this task due to their ability to learn expressive representations through message passing mechanisms \citep{kipf2016semi, hamilton2017inductive, velivckovic2017graph}. Despite their success, the expressive power of GNNs is known to be theoretically bounded by the Weisfeiler–Lehman hierarchy \citep{xu2018powerful}, and their performance often depends heavily on architectural design and training stability.

A key limitation of existing GNN-based approaches is that they rely on end-to-end learning of graph embeddings, which are typically treated as black-box representations. This makes it difficult to interpret what structural information is being utilized and whether the learned representations are inherently informative for the downstream task. Furthermore, training deep graph models can be computationally expensive and sensitive to hyperparameter choices, especially on large-scale or heterogeneous graph datasets.

To mitigate these challenges, several pooling and hierarchical representation methods have been proposed, aiming to construct graph-level embeddings from node representations in a permutation-invariant manner \citep{vinyals2015order, zaheer2017deep, ying2018hierarchical}. While these approaches improve aggregation, they still inherit the limitations of learned embeddings and do not provide explicit mechanisms to evaluate representation quality prior to training.

From a data perspective, benchmark datasets such as TUDataset and Open Graph Benchmark have played a central role in evaluating graph learning methods \citep{morris2020tudataset, hu2020open}. However, performance comparisons on these benchmarks often focus solely on predictive accuracy, without providing insight into the intrinsic informational content of different graph representations.

Recently, the concept of Network Usable Information (NUI) was introduced in \textsc{NetInfoF} \citep{lee2024netinfof}, providing a principled framework for quantifying the amount of task-relevant information contained in learned representations for node-level and link-level tasks. However, extending this notion to graph-level classification remains challenging due to the absence of a unified representation space across graphs with varying sizes and structures.

In this work, we propose \textsc{NetinfoGC}, a framework that extends NUI to graph classification. Our approach constructs a family of permutation-invariant graph representations derived from both propagation-based mechanisms and classical structural descriptors. In particular, we incorporate graph centrality measures as complementary structural signals alongside the original \textsc{NetInfoF} components, enabling a richer characterization of graph topology.

Beyond representation construction, \textsc{NetinfoGC} introduces a two-stage pipeline. First, we estimate NUI using an unsupervised clustering-based procedure that measures alignment between induced structure and ground-truth labels. Second, we exploit the same representations using sparse-group LASSO regularization \citep{simon2013sparse}, which enables automatic selection of informative structural descriptors while suppressing redundant ones.

Graph-based learning has been widely studied in security-related applications such as malware classification \citep{malhotra2024comparison}, smart contract vulnerability detection \citep{zhou2020automating}, botnet detection \citep{zhou2020automating}, and adversarial robustness analysis \citep{pendlebury2019tesseract}. These settings further motivate the need for interpretable and efficient representation evaluation mechanisms, as graph structures evolve dynamically and retraining large models may be impractical.

\textbf{Contributions.} Concretely, this paper makes the following contributions:

\begin{itemize}
    \item We adapt the NUI framework from node/link-level tasks to graph classification, resolving
    the representation-space mismatch that arises when graphs vary in size and structure.
    \item We augment the propagation-based descriptors from NETINFOF with five classical centrality
    measures, and show empirically that these two families capture complementary structural
    information rather than redundant signal.
    \item We design a training-free, clustering-based procedure for estimating NUI at the graph
    level, giving a label-aware but model-free proxy for representation quality.
    \item Empirically, we find that degree centrality is a consistently strong - sometimes the
    strongest - representation across both benchmarks, and that NUI tracks downstream accuracy
    closely enough to serve as a pre-training screening criterion.
\end{itemize}

\section{NetinfoGC for Graph Classification}
\label{sec:method}

This section introduces \textsc{NetinfoGC}, a graph classification framework that extends the Network Usable Information (NUI) paradigm of \textsc{NetInfoF} \citep{lee2024netinfof} to graph-level learning. While the underlying principles of \textsc{NetInfoF} remain unchanged, graph classification requires representations that operate on entire graphs rather than individual nodes or edges. This introduces the need for permutation-invariant, fixed-dimensional graph embeddings that can be compared across graphs of varying sizes and structures.

\subsection{Problem Definition}
\label{subsec:problem_definition_gc}

Let
\[
\mathcal{G}
=
\{(G_i,y_i)\}_{i=1}^N
\]
denote a dataset of $N$ attributed graphs, where each graph
\[
G_i = (V_i, E_i, X_i)
\]
consists of a node set $V_i$, an edge set $E_i$, and a node feature matrix
\[
X_i \in \mathbb{R}^{|V_i| \times f}.
\]
Each graph is associated with a label
\[
y_i \in \{1,\dots,c\},
\]
where $c$ denotes the number of classes.

The objective of graph classification is to learn a function
\[
f : \mathcal{G} \rightarrow \{1,\dots,c\}
\]
that maps unseen graphs to their correct labels. Unlike node-level tasks, graph classification requires learning from heterogeneous graph instances with varying sizes, requiring invariant and fixed-dimensional representations.

\textsc{NetinfoGC} addresses this by first constructing graph-level representations from multiple structural descriptors, and then measuring and exploiting their Network Usable Information in a unified framework.

\subsection{Graph-level Representation Learning}
\label{subsec:graph_representation_learning}

\textsc{NetinfoGC} constructs graph representations from a unified set of structural descriptors that capture complementary aspects of topology, feature propagation, and vertex importance.

\paragraph{Unified descriptor family.}
We define a set of representation functions:
\[
\mathcal{M}
=
\{U, R, F, P, S, Deg, Bet, Close, PR, Eig\},
\]
where each $m \in \mathcal{M}$ denotes either a propagation-based descriptor from \textsc{NetInfoF} or a classical centrality measure.

For each graph $G_i$, each descriptor induces a node-level representation:
\[
Z_i^{(m)} \in \mathbb{R}^{|V_i| \times d}.
\]

\paragraph{Graph-level aggregation.}
To obtain permutation-invariant representations, we apply a READOUT function:
\[
h_i^{(m)} = \mathrm{READOUT}(Z_i^{(m)}) \in \mathbb{R}^{d}.
\]
We use mean pooling as the default READOUT operator, ensuring invariance to node ordering and robustness across graphs of varying size.

\paragraph{Centrality-based representations.}
In addition to propagation-based descriptors, we incorporate classical centrality measures to capture structural importance from a complementary perspective.

For each node $v \in V_i$, we compute:

\[
c^{deg}(v), \quad c^{bet}(v), \quad c^{close}(v), \quad c^{pr}(v), \quad c^{eig}(v).
\]

Each centrality measure produces a vector:
\[
\mathbf{c}_i^{(m)} \in \mathbb{R}^{|V_i|}.
\]

We treat this vector as a node-level feature:
\[
Z_i^{(m)} = \mathbf{c}_i^{(m)},
\]
which is then aggregated using the same READOUT operator:
\[
h_i^{(m)} = \mathrm{READOUT}(Z_i^{(m)}).
\]

\paragraph{Unified graph representation.}
Each graph is represented by the set:
\[
\mathcal{H}_i = \{ h_i^{(m)} \mid m \in \mathcal{M} \}.
\]

The final representation used for downstream tasks is constructed via concatenation:
\[
x_i = \big\|_{m \in \mathcal{M}} h_i^{(m)} \in \mathbb{R}^{|\mathcal{M}| \cdot d}.
\]

This representation integrates propagation-based, feature-based, and centrality-based structural information into a single unified embedding space.

\subsection{NetinfoGC for NUI Measurement}
\label{subsec:netinfogc_probe}

To estimate Network Usable Information (NUI), we evaluate whether a graph representation induces structure that is aligned with the true class labels.

We consider the dataset of graph embeddings:
\[
\mathcal{X} = \{x_i\}_{i=1}^N.
\]

\paragraph{Unsupervised clustering.}
We apply $k$-means clustering to $\mathcal{X}$ to obtain pseudo-labels:
\[
\hat{y}_i = \mathrm{Cluster}(x_i),
\]
where $k \geq c$. Before clustering, embeddings are $\ell_2$-normalized.

\paragraph{NUI estimation.}
We define the conditional entropy:
\[
H(Y \mid \hat{Y}) = - \sum_{y,\hat{y}} p(y,\hat{y}) \log p(y \mid \hat{y}).
\]

The \textsc{NetinfoGC} score is defined as:
\[
\mathrm{NetinfoGC} = 2^{-H(Y \mid \hat{Y})}.
\]

Lower entropy indicates stronger alignment between clustering structure and ground-truth labels, implying higher usable information.

\subsection{NetinfoGC for NUI Exploitation}
\label{subsec:netinfogc_act}

For classification, we concatenate all graph representations:
\[
x_i =
\big\|_{m \in \mathcal{M}} h_i^{(m)}.
\]

We use a logistic regression model with sparse-group LASSO regularization. Let $\beta^{(m)}$ denote parameters associated with descriptor $m$. The optimization problem is:
\[
\min_{\beta}
\mathcal{L}(\beta)
+
\lambda_1 \sum_{m \in \mathcal{M}} \|\beta^{(m)}\|_2
+
\lambda_2 \|\beta\|_1.
\]

This induces:
- group-level sparsity (removal of entire descriptors),
- feature-level sparsity (within descriptors).

Descriptors with low Network Usable Information are automatically suppressed, while informative ones are retained.

Prediction is performed as:
\[
\hat{y}_i = \arg\max_y \sigma(x_i^\top \beta).
\]

\subsection{Computational Complexity}
\label{subsec:complexity}

Representation construction is dominated by propagation operations:
\[
O(k \bar{e} d),
\]
and READOUT:
\[
O(\bar{n} d).
\]

Centrality computation varies:
- degree: $O(\bar{e})$
- betweenness/closeness: $O(\bar{n}\bar{e})$
- PageRank/eigenvector: $O(T\bar{e})$

NUI measurement via $k$-means:
\[
O(T_k N |\mathcal{M}| d).
\]

Sparse-group LASSO training scales linearly:
\[
O(N |\mathcal{M}| d).
\]

Overall complexity remains linear in graph size for sparse graphs.

\subsection{Applications}

\textsc{NetinfoGC} applies to any domain involving graph classification, including molecular property prediction, cybersecurity, program analysis, and communication networks. A key advantage is that it enables estimation of representation quality before training complex graph models.

In cybersecurity, graphs representing program execution or network traffic can be analyzed to determine whether structural descriptors contain sufficient discriminative information. In molecular learning, it allows evaluation of whether structural or centrality-based representations are informative before training GNNs.

Overall, \textsc{NetinfoGC} provides a principled framework for evaluating and exploiting graph representations through Network Usable Information.

\section{Numerical Evaluation}
\label{sec:experiments}

We evaluate \textsc{NetinfoGC} on standard graph classification benchmarks to study (i) the effectiveness of graph-level Network Usable Information (NUI) as a predictor of downstream performance, and (ii) the contribution of different structural descriptors, including both the original \textsc{NetInfoF} components and classical graph centrality measures.

\subsection{Experimental Setup}
\label{subsec:experimental_setup}

We conduct experiments on the \texttt{IMDB-BINARY} and \texttt{IMDB-MULTI} datasets from the TU benchmark suite \citep{morris2020tudataset}. Each graph is converted into a set of graph-level embeddings using the READOUT-based construction described in Section~\ref{subsec:graph_representation_learning}. We evaluate representations derived from both the original \textsc{NetInfoF} components and five centrality-based descriptors: degree, betweenness, closeness, PageRank, and eigenvector centrality.

Unless otherwise specified, we use an embedding dimension of $d=32$, 50 histogram bins, propagation depth $k=32$ for propagation-based descriptors, learning rate $10^{-3}$, and train all models for 500 epochs using sparse-group LASSO regularization. All results are reported as mean $\pm$ standard deviation over multiple independent runs, including NUI score, validation accuracy, and test accuracy.

\subsection{Results}
\label{subsec:results}

Table~\ref{tab:results} summarizes the performance of different graph representations across both datasets. Overall, we observe that structural descriptors exhibit varying degrees of usefulness, and no single representation is universally optimal across datasets.

\paragraph{Effectiveness of centrality-based representations.}
A key observation is that classical centrality measures provide competitive and in some cases superior graph classification performance compared to propagation-based representations. In particular, degree centrality consistently achieves the strongest performance across both datasets, reaching $70.54\%$ test accuracy on \texttt{IMDB-BINARY} and $55.08\%$ on \texttt{IMDB-MULTI}. This is also reflected in its high NUI score, suggesting a strong alignment between the amount of usable structural information and downstream predictive performance.

Other centrality measures show dataset-dependent behavior. PageRank achieves competitive performance on \texttt{IMDB-BINARY}, while betweenness and closeness centralities provide moderate improvements in specific settings. In contrast, eigenvector centrality performs less consistently, particularly on \texttt{IMDB-MULTI}, indicating that global spectral influence alone may not sufficiently capture discriminative graph-level information.

\paragraph{Comparison with \textsc{NetInfoF} components.}
Among the original \textsc{NetInfoF} descriptors, propagation-based and neighborhood-based representations remain competitive but are generally outperformed by the most informative centrality-based representations. This suggests that classical structural measures capture complementary information not fully encoded by propagation-based mechanisms.

\paragraph{Combination of representations.}
We further evaluate a combined representation that aggregates all descriptors. Interestingly, this full combination does not consistently outperform the best individual descriptor. This indicates that simply increasing representation dimensionality does not guarantee improved performance, and may introduce redundancy that dilutes discriminative structure. This observation further motivates the role of NUI as a principled criterion for evaluating representation usefulness prior to model construction.

\paragraph{NUI–performance relationship.}
Across representations, we observe a strong empirical correlation between the estimated NUI score and downstream classification accuracy. Representations with higher NUI tend to yield better predictive performance, supporting the premise that NUI provides a meaningful proxy for task-relevant information in graph representations.

\paragraph{Summary.}
Overall, these results demonstrate that \textsc{NetinfoGC} not only enables effective graph classification using a diverse set of structural descriptors, but also provides a practical mechanism to evaluate the usefulness of such representations before training. The results further highlight that classical graph-theoretic measures can play a significant role in modern graph representation learning when viewed through the lens of Network Usable Information.

\begin{table*}[t]
\centering
\caption{Performance of different graph representations on graph classification benchmarks. Results are reported as mean $\pm$ standard deviation (\%). The best test accuracy for each dataset is shown in bold.}
\label{tab:results}
\small
\begin{tabular}{llccc}
\toprule
Dataset & Representation & Mean NUI & Validation Accuracy & Test Accuracy \\
\midrule

\multirow{9}{*}{IMDB-BINARY}
& Structure & $61.14 \pm 0.83$ & $62.67 \pm 1.42$ & $68.43 \pm 1.41$ \\
& Neighborhood & $60.89 \pm 1.67$ & $68.33 \pm 2.39$ & $66.70 \pm 1.84$ \\
& Propagation (Control) & $63.54 \pm 1.20$ & $64.48 \pm 1.07$ & $68.42 \pm 1.84$ \\
& Degree Centrality & $\mathbf{68.29 \pm 1.31}$ & $\mathbf{75.34 \pm 1.08}$ & $\mathbf{70.54 \pm 1.62}$ \\
& Betweenness Centrality & $60.27 \pm 0.83$ & $67.05 \pm 1.49$ & $69.14 \pm 0.83$ \\
& Closeness Centrality & $59.79 \pm 1.20$ & $66.38 \pm 2.01$ & $62.03 \pm 2.39$ \\
& PageRank Centrality & $63.82 \pm 1.65$ & $62.85 \pm 4.51$ & $70.76 \pm 3.50$ \\
& Eigenvector Centrality & $60.32 \pm 1.15$ & $55.53 \pm 2.86$ & $62.03 \pm 1.38$ \\
& All Components & $55.42 \pm 1.67$ & $66.30 \pm 2.53$ & $65.74 \pm 0.98$ \\

\midrule

\multirow{9}{*}{IMDB-MULTI}
& Structure & $40.38 \pm 0.34$ & $43.03 \pm 6.93$ & $40.30 \pm 0.80$ \\
& Neighborhood & $40.99 \pm 0.53$ & $49.41 \pm 0.44$ & $37.44 \pm 1.55$ \\
& Propagation (Control) & $41.30 \pm 0.59$ & $51.50 \pm 0.73$ & $40.63 \pm 5.71$ \\
& Degree Centrality & $\mathbf{42.85 \pm 0.34}$ & $\mathbf{57.03 \pm 1.30}$ & $\mathbf{55.08 \pm 6.38}$ \\
& Betweenness Centrality & $41.71 \pm 0.74$ & $50.46 \pm 1.45$ & $42.51 \pm 2.64$ \\
& Closeness Centrality & $40.57 \pm 1.00$ & $44.99 \pm 9.25$ & $48.50 \pm 4.61$ \\
& PageRank Centrality & $40.32 \pm 0.54$ & $45.18 \pm 7.22$ & $45.05 \pm 4.79$ \\
& Eigenvector Centrality & $41.22 \pm 0.54$ & $40.63 \pm 5.21$ & $36.20 \pm 3.07$ \\
& All Components & $38.61 \pm 0.64$ & $46.55 \pm 3.18$ & $41.41 \pm 5.79$ \\

\bottomrule
\end{tabular}
\end{table*}

\section{Conclusion}

This work extended the NUI paradigm, originally developed for
node- and link-level tasks in NETINFOF, to the graph classification setting. The main technical
challenge in this extension is the lack of a shared representation space across graphs of varying
size and topology; NETINFOGC addresses this by combining propagation-based descriptors with
classical centrality measures under a common permutation-invariant READOUT operator.

Two findings stand out from the experiments. First, degree centrality alone was consistently
among the strongest representations on both IMDB-BINARY and IMDB-MULTI, often matching or
exceeding the propagation-based descriptors inherited from NETINFOF. This is a useful reminder
that simple, interpretable structural statistics still carry substantial discriminative signal for
graph-level tasks, even relative to learned or propagation-heavy alternatives. Second, the
clustering-based NUI estimate tracked downstream test accuracy fairly closely across
representations, which suggests it can serve as a cheap, label-aware screening step before
committing to a specific representation or training a full model.

One counterintuitive result deserves more scrutiny than we were able to give it here: concatenating
all descriptors together did not outperform the single best descriptor, and in some cases performed
worse. This points to redundancy or noise introduced by naively stacking representations, and
suggests that NUI-guided descriptor selection --- rather than exhaustive concatenation --- may be
the more principled path forward, which is precisely the role the sparse-group LASSO stage is
intended to play.

The current evaluation is limited to two TUDataset benchmarks, and the centrality measures we used
are standard but not exhaustive. Extending NETINFOGC to larger, more heterogeneous graph
collections (e.g., OGB), incorporating attributed or weighted centrality variants, and studying the
theoretical relationship between NUI and generalization error are natural next steps.
\bibliographystyle{apalike}
\bibliography{references}
\end{document}